\documentclass{article}
\usepackage{ijcai18}

\usepackage{times}
\usepackage{xcolor}
\usepackage{soul}
\usepackage[utf8]{inputenc}
\usepackage[small]{caption}

\usepackage{times}
\usepackage{helvet}
\usepackage{courier}
\usepackage{subfigure}
\usepackage{pgfplots}
\usepackage{graphicx}
\usepackage{subfig}
\usepackage{subfloat}
\usepackage{tikz-dependency}
\usepackage{algorithm}
\usepackage{algorithmic}

\usepackage{tabularx}
\usepackage{subfigure}
\usepackage{xparse}
\usepackage{xspace}
\usepackage{relsize}
\usepackage{graphicx}
\usepackage{multirow}
\usepackage[normalem]{ulem}
\usepackage{amsmath}
\usepackage{url}
\usepackage{booktabs}
\usepackage{xcolor}
\usepackage{multirow}
\usepackage{colortbl}

\usepackage{times}  
\usepackage{helvet}  
\usepackage{courier}  
\usepackage{url}  
\usepackage{graphicx}  
 \usepackage{amssymb}

\def\e{\mathbf{e}}

\def\bb{\mathbf{b}}

\def\bW{\mathbf{W}}

\def\JJ{\mathcal{J}}
\allowdisplaybreaks

\DeclareMathOperator*{\softmax}{softmax}
\DeclareMathOperator*{\LSTM}{LSTM}
\DeclareMathOperator*{\KL}{KL}

\newcommand\citet[1]{\citeauthor{#1} [\citeyear{#1}]}
\newcommand{\tabincell}[2]{\begin{tabular}{@{}#1@{}}#2\end{tabular}}

\pgfplotsset{every axis/.append style={
                    axis x line=middle,    
                    axis y line=middle,    
                    axis line style={->}, 
                    xlabel={$x$},          
                    ylabel={$y$},          
                    label style={font=\tiny},
                    tick label style={font=\tiny},
                    legend style={font=\tiny},
                    legend pos=outer north east
                    }}

\tikzset{>=stealth}

\newenvironment{itemize*}%
  {\begin{itemize}%
    \setlength{\itemsep}{1pt}%
    \setlength{\parskip}{0pt}}%
  {\end{itemize}}
  \newenvironment{enumerate*}%
  {\begin{enumerate}%
    \setlength{\itemsep}{0pt}%
    \setlength{\parskip}{0pt}}%
  {\end{enumerate}}

\title{Toward Diverse Text Generation with Inverse Reinforcement Learning}

\author{Zhan Shi, Xinchi Chen, Xipeng Qiu\thanks{Corresponding Author, xpqiu@fudan.edu.cn}, Xuanjing Huang\\
 Shanghai Key Laboratory of Intelligent Information Processing, Fudan University\\
School of Computer Science, Fudan University\\
}

\begin{document}

\maketitle

\begin{abstract}
Text generation is a crucial task in NLP. Recently, several adversarial generative models have been proposed to improve the exposure bias problem in text generation. Though these models gain great success, they still suffer from the problems of reward sparsity and mode collapse. In order to address these two problems, in this paper, we employ inverse reinforcement learning (IRL) for text generation. Specifically, the IRL framework learns a reward function on training data, and then an optimal policy to maximum the expected total reward. Similar to the adversarial models, the reward and policy function in IRL are optimized alternately. Our method has two advantages: (1)  the reward function can produce more dense reward signals. (2) the generation policy, trained by ``entropy regularized'' policy gradient,  encourages to generate more diversified texts. Experiment results demonstrate that our proposed method can generate higher quality texts than the previous methods.
\end{abstract}

\section{Introduction}
Text generation is one of the most attractive problems in NLP community. It has been widely used in machine translation, image captioning, text summarization and dialogue systems.

Currently, most of the existing methods \cite{graves2013generating} adopt auto-regressive models to predict the next words based on the historical predictions. Benefiting from the strong ability of deep neural models, such as long short-term memory (LSTM) \cite{hochreiter1997long}, these auto-regressive models can achieve excellent performance. However, they suffer from the so-called \textit{exposure bias} issue \cite{bengio2015scheduled} due to the discrepancy distribution of histories between the training and inference stage. In training stage, the model predicts the next word according to ground-truth histories from the data distribution rather than its own historical predictions from the model distribution.

Recently, some methods have been proposed to alleviate this problem, such as scheduled sampling \cite{bengio2015scheduled}, Gibbs sampling \cite{su2018incorporating} and adversarial models, including SeqGAN \cite{yu2017seqgan}, RankGAN \cite{lin2017adversarial}, MaliGAN \cite{che2017maximum} and LeakGAN \cite{guo2017long}.
Following the framework of generative adversarial networks (GAN) \cite{goodfellow2014generative}, the adversarial text generation models use a discriminator to judge whether a given text is real or not. Then a generator is learned to maximize the reward signal provided by the discriminator via reinforcement learning (RL). Since the generator always generates a entire text sequence, these adversarial models can avoid the problem of exposure bias.

Inspired of their success, there are still two challenges in the adversarial model.

The first problem is \textit{reward sparsity}. The adversarial model depends on the ability of the discriminator, therefore we wish the discriminator always correctly discriminates the real texts from the ``generated'' ones. Instead, a perfect discriminator increases the training difficulty due to the sparsity of the reward signals.
There are two kinds of work to address this issue. The first one is to improve the signal from the discriminator. RankGAN \cite{lin2017adversarial} uses a ranker to take place of the discriminator, which can learn the relative ranking information between the generated and the real texts in the adversarial framework. MaliGAN \cite{che2017maximum} develops normalized maximum likelihood optimization target to alleviate the reward instability problem. The second one is to decompose the discrete reward signal into various sub-signals. LeakGAN \cite{guo2017long} takes a hierarchical generator, and in each step, generates a word using leaked information from the discriminator.

The second problem is the \textit{mode collapse}. The adversarial model tends to learn limited patterns because of mode collapse. One kind of methods, such as TextGAN \cite{zhang2017adversarial}, uses feature matching \cite{salimans2016improved,metz2016unrolled} to alleviate this problem, it is still hard to train due to the intrinsic nature of GAN.
Another kind of methods \cite{bayer2014learning,chung2015recurrent,serban2017hierarchical,wang2017text} introduces latent random variables to model the variability of the generated sequences.



To tackle these two challenges, we propose a new method to generate diverse text via inverse reinforcement learning (IRL) \cite{ziebart2008maximum}. Typically, the text generation can be regarded as an IRL problem. Each text in the training data is generated by some experts with an unknown reward function. There are two alternately steps in IRL framework. Firstly, a reward function is learned to explain the expert behavior. Secondly, a generation policy is learned to maximize the expected total rewards.
The reward function aims to increase the rewards of the real texts in training set and decrease the rewards of the generated texts. Intuitively, the reward function plays the similar role as the discriminator in SeqGAN. Unlike SeqGAN, the reward function is an instant reward of each step and action, thereby providing more dense reward signals.
The generation policy generates text sequence by sampling one word at a time. The optimized policy be learned by ``entropy regularized'' policy gradient \cite{finn2016guided}, which  intrinsically leads to a more diversified text generator.

The contributions of this paper are summarized as follows.
\begin{itemize*}
  \item We regard text generation as an IRL problem, which is a new perspective on this task.
  \item Following the maximum entropy IRL \cite{ziebart2008maximum}, our method can improve the problems of reward sparsity and mode collapse.
  \item To better evaluate the quality of the generated texts, we propose three new metrics based on BLEU score, which is very similar to precision, recall and $F_1$ in traditional machine learning task.
\end{itemize*}


\section{Text Generation via Inverse Reinforcement Learning}
\begin{figure}[t]
  \centering
  \includegraphics[width=0.37\textwidth]{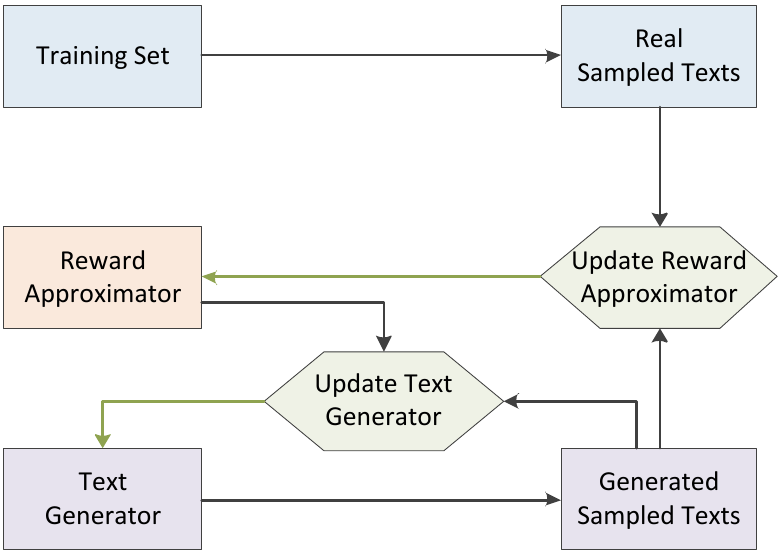}
  \caption{IRL framework for text generation.}\label{fig:IRL_diagram}
\end{figure}

Text generation is to generate a text sequence $x_{1:T}=x_1,x_2,\cdots,x_T$ with a parameterized auto-regressive probabilistic model $q_\theta(x)$, where $x_t$ is a word in a given vocabulary $\mathcal{V}$. The generation model $q_\theta(x)$ is learned from a given dataset $\{x^{(n)}\}_{n=1}^N$ with an underlying generating distribution $p_{data}$.

In this paper, we formulate text generation as inverse reinforcement learning (IRL) problem.
Firstly, the process of text generation can be regarded as Markov decision process (MDP). In each timestep $t$, the model generates $x_t$ according a policy $\pi_\theta(a_t|s_t)$, where $s_t$ is the current state of the previous prediction $x_{1:t}$ and $a_t$ is the action to select the next word $x_{t+1}$.
A text sequence $x_{1:T}=x_1,x_2,\cdots,x_T$  can be formulated by a trajectory of MDP
$\tau= \{s_1,a_1,s_2,a_2...,s_T,a_T\}$. Therefore, the probability of $x_{1:T}$ is
\begin{equation}
q_\theta(x_{1:T}) = q_\theta(\tau) =\prod_{t=1}^{T-1} \pi_\theta(a_t=x_{t+1}|s_t=x_{1:t}),
\end{equation}
where the state transition $p(s_{t+1}=x_{1:t+1}|s_t=x_{1:t},a_t=x_{t+1})=1$ is deterministic and can be ignored.

Secondly, the reward function is not explicitly given for text generation. Each text sequence $x_{1:T}=x_1,x_2,\cdots,x_T$ in the training dataset is formulated by a trajectory $\tau$ by experts from the distribution $p(\tau)$, and we have to learn a reward function that explains the expert behavior.

Concretely, IRL consists of two phases: (1) estimate the underlying reward function of experts from the training dataset; (2) learn an optimal policy to generate texts, which aims to maximize the expected rewards.
These two phases are  executed alternately. The framework of our method is as shown in Figure \ref{fig:IRL_diagram}.


\subsection{Reward Approximator}
\begin{figure}[t]
  \centering
  \includegraphics[width=0.40\textwidth]{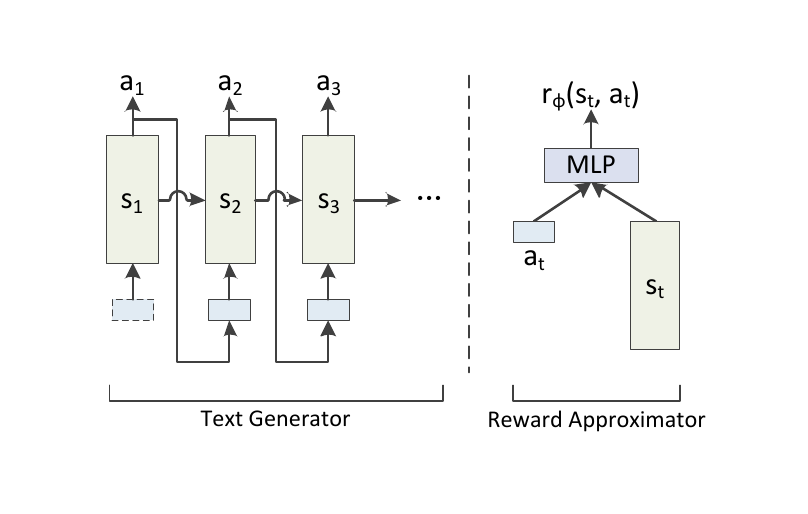}
  \caption{Illustration of text generator and reward approximator.}\label{fig:IRL_model}
\end{figure}

Following the framework of maximum entropy IRL \cite{ziebart2008maximum}, we assume that the texts in training set are sampled from the
distribution $p_\phi(\tau)$,
\begin{equation}
p_\phi(\tau) = \frac{1}{Z} \exp(R_\phi(\tau)), \label{eq:ptau}
\end{equation}
where $R_\phi(\tau)$ an unknown reward function parameterized by $\phi$, $Z = \int_\tau \exp(R_\phi(\tau)) d\tau$ is the partition function.

The reward of trajectory $R_\phi(\tau)$ is a parameterized reward function and assumed to be summation of the rewards of each steps $r_\phi(s_t,a_t)$:
\begin{equation}
R_\phi(\tau) = \sum_t r_\phi(s_t,a_t),
\end{equation}
where $r_\phi(s_t,a_t)$ is modeled a simple feed-forward neural network as shown in Figure \ref{fig:IRL_model}.

\subsubsection{Objective of Reward Approximator}

The objective of the reward approximator is to maximize the log-likehood of the samples in the training set:
\begin{equation}
\JJ_r(\phi) = \frac{1}{N} \sum_{n = 1}^{N} \log p_\phi(\tau_{n}) =  \frac{1}{N}\sum_{n = 1}^{N} R_\phi(\tau_{n}) - \log Z, \label{eq:max-loss-one}
\end{equation}
where $\tau_{n}$ denotes the $n_{th}$ sample in the training set $D_{train}$.

Thus, the derivative of $\JJ_r(\phi)$ is:
\begin{align}
\nabla_\phi \JJ_r(\phi) &\!= \!\frac{1}{N}\!\! \sum_{n} \!\nabla_\phi R_\phi(\tau_{n}) \!-\! \frac{1}{Z}\!\! \int_\tau\!\!
\exp(R_\phi(\tau))\nabla_\phi R_\phi(\tau) \mathrm{d}\tau \nonumber\\
&\!=\! \mathbb{E}_{\tau \sim p_{data}} \nabla_\phi R_\phi(\tau) - \mathbb{E}_{\tau \sim p_\phi(\tau)}\nabla_\phi R_\phi(\tau).\label{eq:de_loss_r}
\end{align}
Intuitively, the reward approximator aims to increase the rewards of the real texts and decrease the trajectories drawn from the distribution $p_\phi(\tau)$. As a result, $p_\phi(\tau)$ will be an approximation of $p_{data}$.

\paragraph{Importance Sampling}
Though it is quite straightforward to sample $\tau \sim p_\phi(\tau)$ in Eq. (\ref{eq:de_loss_r}), it is actually inefficient in practice. Instead, we directly use trajectories sampled by text generator $q_\theta(\tau)$ with importance sampling. Concretely, Eq. (\ref{eq:de_loss_r}) is now formalized as:
{\small\begin{align}
\nabla_\phi \JJ_r(\phi) 
\!&\approx\! \frac{1}{N} \! \sum_{i=1}^N \! \nabla_\phi R_\phi(\tau_{i}) \!-\! \frac{1}{\sum_j{w_j}} \! \sum_{j=1}^{M} \! w_j \nabla_\phi R_\phi(\tau_{j}^\prime),\label{eq:de_importance_sampling}
\end{align}}
where $w_j \propto \frac{\exp(R_\phi(\tau_{j}))}{q_{\theta}(\tau_j)}$. For each batch, we sample $N$ texts from the train set and $M$ texts drawn from $q_\theta$.

\subsection{Text Generator}
The text generator uses a policy $\pi_\theta(a|s)$ to predict the next word one by one. The current state $s_t$ can be modeled by LSTM
neural network as shown in Figure \ref{fig:IRL_model}. For $\tau= \{s_1,a_1,s_2,a_2...,s_T,a_T\}$,
\begin{gather}
\mathbf{s}_t = \LSTM(\mathbf{s}_{t-1}, \e_{a_{t-1}}),\\
\pi_\theta(\mathbf{a}_t|s_t) = \softmax(\bW \mathbf{s}_t + \bb),
\end{gather}
where $\mathbf{s}_t$ is the vector representation of state $s_t$; $\mathbf{a}_t$ is distribution over the vocabulary;
 $\e_{a_{t-1}}$ is the word embedding of $a_{t-1}$; $\theta$ denotes learnable parameters including $\bW$, $\bb$ and all the parameters of LSTM.

\subsubsection{Objective of Text Generator}
Following ``entropy regularized'' policy gradient \cite{williams1992simple,nachum2017bridging}, the objective of text generator is to  maximize the expected reward plus an entropy regularization.
\begin{align}
\JJ_g(\theta) =\mathbb{E}_{\tau \sim q_\theta(\tau)} [R_\phi(\tau)]+ H(q_\theta(\tau))
\end{align}
where $H(q_\theta(\tau)) = -\mathbb{E}_{q_\theta(\tau)}[\log q_\theta(\tau)]$ is an entropy term, which can prevent premature entropy collapse and encourage the policy to generate more diverse texts.

Intuitively, the ``entropy regularized'' expected reward can be rewrite as
\begin{align}
\JJ_g(\theta) &= -\KL(q_\theta(\tau)||p_\phi(\tau)) + \log Z,
\end{align}
where $Z = \int_\tau \exp(R_\phi(\tau)) d\tau$ is the partition function and can be regarded as a constant unrelated to $\theta$.
Therefore, the objective is also to minimize the KL divergence between the text generator $q_\theta(\tau)$ and the underlying distribution $p_\phi(\tau)$.

\begin{figure}[t]
  \centering
  \includegraphics[width=0.30\textwidth]{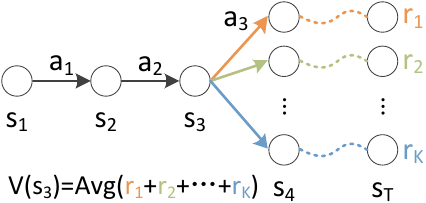}
  \caption{MCMC sampling for calculating the expected total reward at each state.}\label{fig:MCMC}
\end{figure}

Thus, the derivative of $\JJ_g(\theta)$ is
\begin{align}
\nabla_\theta \JJ_g(\theta) =& \sum_t \mathbb{E}_{\pi_\theta(a_t|s_t)} \nabla_\theta \log\pi_{\theta}(a_t|s_t) \nonumber\\
&\left[R_\phi(\tau_{t:T})-  \log \pi_\theta (a_{t}|s_{t})-1\right].\label{eq:de_loss_g}
\end{align}
where $R_\phi(\tau_{t:T})$ denotes the reward of partial trajectory $\tau_t,\cdots,\tau_T$. For obtaining lower variance, $R(\tau_{t:T})$ can be approximately computed by
\begin{equation}
R_\phi(\tau_{t:T}) \approx r_\phi(s_{t}, a_{t}) + V(s_{t+1}),
\end{equation}
where $V(s_{t+1})$ denotes the expected total reward at state $s_{t+1}$ and can be approximately computed by MCMC. Figure \ref{fig:MCMC} gives an illustration.

\subsection{Why Can IRL  Alleviate Mode Collapse?}
GANs often suffer from mode collapse, which is partially caused by the use of Jensen-Shannon (JS) divergence. There is a reverse KL divergence $\KL(q_\theta(\tau)\|p_{data})$ in JS divergence. Since the $p_{data}$ is approximated by training data, the reverse KL divergence encourages $q_\theta(\tau)$ to generate safe samples and avoid generating samples where the training data does not occur. In our method, the objective is $\KL(q_\theta(\tau)||p_\phi(\tau))$. Different from GANs, we use $p_\phi(\tau)$ in IRL framework instead of $p_{data}$. Since $p_\phi(\tau)$ never equals to zero due to its assumption, IRL can alleviate the model collapse problem in GANs.

\section{Training}

The training procedure consists of two steps: (I) reward approximator update step (\textbf{r-step}) and (II) text generator update step (\textbf{g-step}). These two steps are applied iteratively as described in Algorithm (\ref{alg:inverse-rl}).

Initially, we have $r_\phi$ with random parameters and $\pi_\theta$ with pre-trained parameters by maximum log-likelihood estimation on $D_{train}$. The r-step aims to update $r_\phi$ with $\pi_\theta$ fixed. The g-step aims to update $\pi_\theta$ with $r_\phi$ fixed.

\begin{algorithm}[t]
\begin{algorithmic}[1]
\REPEAT
    \STATE Pretrain $\pi_\theta$ on $D_{train}$ with MLE
	\FOR {$n_r$ epochs in r-step}
        \STATE Drawn $\tau^{(1)}, \tau^{(2)}, \cdots, \tau^{(i)}, \cdots, \tau^{(N)} \sim p_{data}$
    	\STATE Drawn $\tau^{\prime(1)}, \tau^{\prime(2)}, \cdots, \tau^{\prime(j)}, \cdots, \tau^{\prime(M)} \sim q_\theta$
        \STATE Update $\phi \leftarrow \phi + \alpha \nabla_\phi \JJ_r(\phi)$
    \ENDFOR
    \FOR {$n_g$ batches in g-step}
    	\STATE Drawn $\tau^{(1)}, \tau^{(2)}, \cdots, \tau^{(i)}, \cdots, \tau^{(N)} \sim q_\theta$
        \STATE Calculate expected reward $R_\phi(\tau_{t:T})$ by MCMC
        \STATE Update $\theta \leftarrow \theta + \beta \nabla_\theta \JJ_g(\theta)$
    \ENDFOR
\UNTIL{Convergence}
\end{algorithmic}
\caption{\textbf{IRL for Text Generation}}
\label{alg:inverse-rl}
\end{algorithm}

\section{Experiment}
To evaluate the proposed model, we experiment on three corpora: the synthetic oracle dataset \cite{yu2017seqgan}, the COCO image caption dataset \cite{chen2015microsoft} and the IMDB movie review dataset \cite{diao2014jointly}. Furthermore, we also evaluate the performance by human on the image caption dataset and the IMDB corpus. Experimental results show that Our method outperforms the previous methods. Table \ref{tab:Parameter-setting} gives the experimental settings on the three corpora. 

\begin{table}[t]\setlength{\tabcolsep}{0.5pt}\small
\centering
\begin{tabular}{lccc} \toprule
\multirow{2}*{Hyper-Parameters} & \multicolumn{2}{c}{Synthetic Oracle} & \multirow{2}*{COCO \& IMDB}\\
&L = 20&L = 40&\\
\midrule
\textbf{Text Generator}   \\
- Embedding dimension & 32 & 64 & 128 \\
- Hidden layer dimension & 32 & 64 & 128 \\
- Batch size  & \multicolumn{2}{c}{64} & 128 \\
- Optimizer \& lr rate  & \multicolumn{2}{c}{Adam, 0.005} & Adam, 0.005\\
\midrule
\textbf{Reward Approximator} \\
- Drop out & 0.75 & 0.45 & 0.75 \\
- Batch size & \multicolumn{2}{c}{64} & 1024 \\
- Optimizer \& lr rate & \multicolumn{2}{c}{Adam, 0.0004} & Adam, 0.0004 \\ \bottomrule
\end{tabular}
\caption{Configurations on hyper-parameters.}
\label{tab:Parameter-setting}
\end{table}

\begin{table}[t]\setlength{\tabcolsep}{2pt}\small
\centering
\begin{tabular}{c|ccccc||c} \toprule
Length & MLE & SeqGAN & RankGAN & LeakGAN & IRL & \tabincell{c}{Ground\\ Truth} \\ \midrule
20 & 9.038$^*$ & 8.736$^*$ & 8.247$^*$ & 7.038$^*$ & \textbf{6.913} & 5.750 \\
40 & 10.411$^*$ & 10.310$^*$ & 9.958$^*$ & 7.197$^*$ & \textbf{7.083} & 4.071 \\ \bottomrule
\end{tabular}
\caption{The overall NLL performance on synthetic data. ``Ground Truth'' consists of samples generated by the oracle LSTM model. Results with * are reported in their papers.}
\label{tab:synthetic}
\end{table}

\begin{figure}[!t] \tiny 
  \centering
  \pgfplotsset{width=0.29\textwidth}
  \ref{learning_curves}\\
\begin{tikzpicture}
\node[draw] at (2.5,2.5) {Text Length = 20};
    \begin{axis}[
    xlabel={Learning epochs},
   ylabel={NLL loss},
   xmin=0,xmax=250,
        ymin=6.5,
    mark size=0.5pt,
    ymajorgrids=true,
    grid style=dashed,
    legend columns=-1,
    legend entries={10 Epochs, 25 Epochs, 50 Epochs, 100 Epochs},
    legend style={/tikz/every even column/.append style={column sep=0.13cm}},
    legend to name=learning_curves,
    ]
    \addplot [black, dashed, mark=square*] table [x index=0, y index=4] {IRL_pretrain.txt};
    \addplot [blue, dashed, mark=*] table [x index=0, y index=1] {IRL_pretrain.txt};
    \addplot [red] table [x index=0, y index=2] {IRL_pretrain.txt};
    \addplot [green, dashed] table [x index=0, y index=3] {IRL_pretrain.txt};
    \end{axis}
\end{tikzpicture}
\hspace{-0.8em}
\begin{tikzpicture}
\node[draw] at (2.5,2.5) {Text Length = 40};
    \begin{axis}[
    xlabel={Learning epochs},
    ylabel={NLL loss},
    xmin=0,xmax=250,
        ymin=6.5,
    mark size=0.5pt,
    ymajorgrids=true,
    grid style=dashed,
    ]
    \addplot [black, dashed, mark=square*] table [x index=0, y index=1] {IRL_pretrain40.txt};
    \addplot [blue, dashed, mark=*] table [x index=0, y index=2] {IRL_pretrain40.txt};
    \addplot [red] table [x index=0, y index=3] {IRL_pretrain40.txt};
    \addplot [green, dashed] table [x index=0, y index=4] {IRL_pretrain40.txt};
    \end{axis}
\end{tikzpicture}
\caption{Learning curves with different pretrain epochs (10, 25, 50, 100 respectively) on texts of length 20 and 40.}
\label{fig:sythetic-curve-pretrain}

\qquad

\centering
  \pgfplotsset{width=0.29\textwidth}
  \ref{learning_balance}\\
\begin{tikzpicture}
\node[draw] at (2.5,2.5) {Text Length = 20};
    \begin{axis}[
    xlabel={Learning epochs},
   ylabel={NLL loss},
   xmin=0,xmax=250,
        ymin=6.5,
    mark size=0.4pt,
    ymajorgrids=true,
    grid style=dashed,
    legend columns=-1,
    legend entries={1:10, 1:1, 10:1},
    legend style={/tikz/every even column/.append style={column sep=0.13cm}},
    legend to name=learning_balance,
    ]
    \addplot [black, dashed, mark=*] table [x index=0, y index=3] {IRL_equilibrium.txt};
    \addplot [blue, dashed] table [x index=0, y index=1] {IRL_equilibrium.txt};
    \addplot [red] table [x index=0, y index=2] {IRL_pretrain.txt};
    \end{axis}
\end{tikzpicture}
\hspace{-0.8em}
\begin{tikzpicture}
\node[draw] at (2.5,2.5) {Text Length = 40};
    \begin{axis}[
    xlabel={Learning epochs},
    ylabel={NLL loss},
    xmin=0,xmax=250,
        ymin=6.5,
    mark size=0.4pt,
    ymajorgrids=true,
    grid style=dashed,
    ]
    \addplot [black,dashed, mark=*] table [x index=0, y index=2] {IRL_equilibrium40.txt};
    \addplot [blue, dashed] table [x index=0, y index=1] {IRL_equilibrium40.txt};
    \addplot [red] table [x index=0, y index=1] {IRL_pretrain40.txt};
    \end{axis}
\end{tikzpicture}
\caption{Learning curves with different training equilibriums between text generator and reward approximator on texts of length 20 and 40. The proportion in the legend means $n_r : n_g$.}
\label{fig:sythetic-curve-equilibrium}

\qquad

  \centering
  \pgfplotsset{width=0.29\textwidth}
  \ref{different_model}\\
\begin{tikzpicture}
\node[draw] at (2.5,2.5) {Text Length = 20};
    \begin{axis}[
    xlabel={Learning epochs},
   ylabel={NLL loss},
   xmin=0,xmax=250,
        ymin=6.5,
    mark size=0.5pt,
    ymajorgrids=true,
    grid style=dashed,
    legend columns=-1,
    legend entries={IRL,LeakGAN,SeqGAN,MLE},
    legend style={/tikz/every even column/.append style={column sep=0.13cm}},
    legend to name=different_model,
    ]
    \addplot [red] table [x index=0, y index=2] {IRL_pretrain.txt};
    \addplot [black, dashed, mark=square*] table [x index=0, y index=1] {leakgan20.txt};
    \addplot [blue, dashed, mark=*] table [x index=0, y index=1] {seqgan20.txt};
    \addplot [green, dashed] table [x index=0, y index=1] {mle20.txt};
    \addplot [black, dashed] table [x index=0, y index=1] {vertical20.txt};
    \end{axis}
\end{tikzpicture}
\hspace{-0.5em}
\begin{tikzpicture}
\node[draw] at (2.5,2.5) {Text Length = 40};
    \begin{axis}[
    xlabel={Learning epochs},
    ylabel={NLL loss},
    xmin=0,xmax=250,
        ymin=6.5,
    mark size=0.5pt,
    ymajorgrids=true,
    grid style=dashed,
    ]
    \addplot [red] table [x index=0, y index=1] {IRL_pretrain40.txt};
    \addplot [black, dashed, mark=square*] table [x index=0, y index=1] {leakgan40.txt};
    \addplot [blue, dashed, mark=*] table [x index=0, y index=1] {seqgan40.txt};
    \addplot [green, dashed] table [x index=0, y index=1] {mle40.txt};
    \addplot [black, dashed] table [x index=0, y index=1] {vertical40.txt};
    \end{axis}
\end{tikzpicture}
\caption{Learning curves of different methods on the synthetic data of length 20 and 40 respectively.The vertical dashed line indicates the end of the pre-training of SeqGAN, LeakGAN and our method respectively. Since RankGAN didn't publish code, we cannot plot the result of RankGAN.
}
\label{fig:sythetic-curve}
\end{figure}
\subsection{Synthetic Oracle}
 The synthetic oracle dataset is a set of sequential tokens which are regraded as simulated data comparing to the real-world language data. It uses a randomly initialized LSTM \footnote{The synthetic data and the oracle LSTM are publicly available at https://github.com/LantaoYu/SeqGAN and https://github.com/CR-Gjx/LeakGAN} as the oracle model to generate 10000 samples of length 20 and 40 respectively as the training set for the following experiments.

The oracle model, which has an intrinsic data distribution $P_{oracle}$, can be used to evaluate the sentences generated by the generative models. The average negative log-likelihood(NLL) is usually conducted to score the quality of the generated sequences \cite{yu2017seqgan,guo2017long,lin2017adversarial}. The lower the NLL score is, the better token sequences we have generated.


\paragraph{Training Strategy}
In experiments, we find that the stability and performance of our framework depend on the training strategy. Figure \ref{fig:sythetic-curve-pretrain} shows the effects of pretraining epochs. It works best in generating texts of length 20 with 50 epochs of MLE pretraining, and in generating texts of length 40 with 10 epochs of pretraining.

Figure \ref{fig:sythetic-curve-equilibrium} shows that the proportion of $n_r : n_g$ in Algorithm \ref{alg:inverse-rl} affects the convergence and final performance. It implies that sufficient training on the approximator in each iteration will lead to better results and convergence. Therefore, we take $n_r : n_g = 10 : 1$ as our final training configuration.

\paragraph{Results} Table \ref{tab:synthetic} gives the results. We compare our method with other previous state-of-the-art methods: maximum likelihood estimation (MLE), SeqGAN, RankGAN and LeakGAN. The listed ground truth values are the average NLL of the training set. Our method outperforms the previous state-of-the-art results (6.913 and 7.083 on length of 20 and 40 respectively).  Figure \ref{fig:sythetic-curve} shows that Our method convergences faster and obtains better performance than other state-of-art methods. 

\paragraph{Analysis} Our method performs better due to the instant rewards approximated at each step of generation. It addresses the reward sparsity issue occurred in previous methods. Thus, the dense learning signals guide the generative policy to capture the underlying distribution of the training data more efficiently.

\subsection{COCO Image Captions}
The image caption dataset \cite{chen2015microsoft} consists of image-description pairs. The length of captions is between 8 and 20. Following LeakGAN \cite{guo2017long}, for preprocessing, we remove low frequency words (less than 10 times) as well as the sentences containing them. We randomly choose 80,000 texts as training set, and another 5,000 as test set.  The vocabulary size of the dataset is 4,939. The average sentence length is 12.8.

\paragraph{New Evaluation Measures on BLEU} To evaluate different methods, we employ BLEU score to evaluate the qualities of the generated texts.
\begin{itemize}
  \item \textit{Forward BLEU (BLEU$_\text{F}$)} uses the testset as reference, and evaluates each generated text with BLEU score.
  \item \textit{Backward BLEU (BLEU$_\text{B}$)} uses the generated texts as reference, and evaluates each text in testset with BLEU score.
  \item \textit{BLEU$_\text{HA}$} is the harmonic average value of BLEU$_\text{F}$ and BLEU$_\text{B}$.
\end{itemize}
Intuitively, BLEU$_\text{F}$ aims to measure the precision (quality) of the generator, while BLEU$_\text{B}$ aims to measure the recall (diversity) of the generator.

The configurations of three proposed valuation measures are shown in Table \ref{tab:diff-evaluate-metric}.

\begin{table}[h]\setlength{\tabcolsep}{4pt}\small
\centering
\begin{tabular}{c|c|c} \toprule
Metrics & Evaluated Texts & Reference Texts \\ \midrule
BLEU$_\text{F}$ & Generated Texts & Test Set \\
BLEU$_\text{B}$ & Test Set & Generated Texts \\
\midrule
BLEU$_\text{HA}$&\multicolumn{2}{c}{$\frac{2\times\text{BLEU}_\text{F}\times\text{BLEU}_\text{B}}{\text{BLEU}_\text{F}+\text{BLEU}_\text{B}}$}\\
\bottomrule
\end{tabular}
\caption{Configurations of BLEU$_\text{F}$, BLEU$_\text{B}$ and BLEU$_\text{HA}$.}
\label{tab:diff-evaluate-metric}
\end{table}

\begin{table}[t]\setlength{\tabcolsep}{1.5pt}\small
\centering
\begin{tabular}{c|  ccccc||c} \toprule
Metrics & MLE & SeqGAN & RankGAN & LeakGAN & IRL & \tabincell{c}{Ground\\Truth} \\ \midrule
BLEU$_\text{F}$-2 & 0.798 & 0.821 & 0.850$^*$ & \textbf{0.914} & 0.829 & 0.836 \\
BLEU$_\text{F}$-3 & 0.631 & 0.632 & 0.672$^*$ & \textbf{0.816} & 0.662 & 0.672 \\
BLEU$_\text{F}$-4 & 0.498 & 0.511 & 0.557$^*$ & \textbf{0.699} & 0.586 & 0.598 \\
BLEU$_\text{F}$-5 & 0.434 & 0.439 & 0.544$^*$ & \textbf{0.632} & 0.542 & 0.557 \\ \midrule

BLEU$_\text{B}$-2 & 0.801 & 0.682 & - & 0.790 & \textbf{0.868} & 0.869 \\
BLEU$_\text{B}$-3 & 0.622 & 0.542 & - & 0.605 & \textbf{0.718} & 0.710 \\
BLEU$_\text{B}$-4 & 0.551 & 0.513 & - & 0.549 & \textbf{0.660} & 0.649 \\
BLEU$_\text{B}$-5 & 0.508 & 0.469 & - & 0.506 & \textbf{0.609} & 0.601 \\ \midrule

BLEU$_\text{HA}$-2 & 0.799 & 0.745 & - &0.847 & \textbf{0.848} & 0.852 \\
BLEU$_\text{HA}$-3 & 0.626 & 0.584 & - &\textbf{0.695} & 0.689 & 0.690 \\
BLEU$_\text{HA}$-4 & 0.523 & 0.512 & - & 0.615 & \textbf{0.621} & 0.622 \\
BLEU$_\text{HA}$-5 & 0.468 & 0.454 & - & 0.562 & \textbf{0.574} & 0.578 \\ \bottomrule
\end{tabular}
\caption{Results on COCO image caption dataset. Results of RankGAN with * are reported in \protect\cite{guo2017long}. Results of MLE, SeqGAN and LeakGAN are based on their published implementations.}
\label{tab:BLEU-COCO}
\end{table}
\paragraph{BLEU$_\text{F}$} 
For BLEU$_\text{F}$, we sample 1000 texts for each method as evaluated texts. The reference texts are the whole test set. We list the BLEU$_\text{F}$ scores of different frameworks and ground truth as shown in first subtable of Table \ref{tab:BLEU-COCO}. Surprisingly, it shows that results of LeakGAN beat the rest, even the ground truth (LeakGAN has averagely 10 points higher than the ground truth). It may due to the mode collapse which frequently occurs in GAN. The text generator is prone to generate safe text patterns but misses many other patterns. Therefore, BLEU$_\text{F}$ is failing to measure the diversity of the generated sentences. 

\paragraph{BLEU$_\text{B}$} 
For BLEU$_\text{B}$, we sample 5000 texts for each method as reference texts. The evaluated texts consist 1000 texts sampled from the test set. The BLEU$_\text{B}$ of each method is listed in the second block of Table \ref{tab:BLEU-COCO}. Intuitively, the higher the BLEU$_\text{B}$ score is, the more diversity the generator gets. From Table  \ref{tab:BLEU-COCO}, our method outperforms the other methods, which implies that our method generates more diversified texts than the other methods. As we have analyzed before, the diversity of our method may be derived from  ``entropy regularization'' policy gradient.

\paragraph{BLEU$_\text{HA}$} 
Finally, BLEU$_\text{HA}$ takes both generation quality and diversity into account and the results are shown in the last block of Table \ref{tab:BLEU-COCO}. The BLEU$_\text{HA}$ reveals that our work gains better performance than other methods.

\begin{table}[t]\setlength{\tabcolsep}{4pt}\small
\centering
\begin{tabular}{c|  cccc||c} \toprule
Metrics & MLE & SeqGAN & LeakGAN & IRL & \tabincell{c}{Ground\\Truth} \\ \midrule
BLEU$_\text{F}$-2 & 0.652 & 0.683 & \textbf{0.809} & 0.788 & 0.791 \\
BLEU$_\text{F}$-3 & 0.405 & 0.418 & \textbf{0.554} & 0.534 & 0.539 \\
BLEU$_\text{F}$-4 & 0.304 & 0.315 & \textbf{0.358} & 0.352 & 0.355 \\
BLEU$_\text{F}$-5 & 0.202 & 0.221 & 0.252 & \textbf{0.262} & 0.258 \\ \midrule

BLEU$_\text{B}$-2 & 0.672 & 0.615 & 0.730 & \textbf{0.755} & 0.785 \\
BLEU$_\text{B}$-3 & 0.495 & 0.451 & 0.483 & \textbf{0.531} & 0.534 \\
BLEU$_\text{B}$-4 & 0.316 & 0.299 & 0.318 & \textbf{0.347} & 0.357 \\
BLEU$_\text{B}$-5 & 0.226 & 0.209 & 0.232 & \textbf{0.254} & 0.258 \\ \midrule

BLEU$_\text{HA}$-2 & 0.662 & 0.647 & 0.767 & \textbf{0.771} & 0.788 \\
BLEU$_\text{HA}$-3 & 0.445 & 0.434 & 0.516 & \textbf{0.533} & 0.537 \\
BLEU$_\text{HA}$-4 & 0.310 & 0.307 & 0.337 & \textbf{0.350} & 0.356 \\
BLEU$_\text{HA}$-5 & 0.213 & 0.215 & 0.242 & \textbf{0.258} & 0.258 \\ \bottomrule
\end{tabular}
\caption{Results on IMDB Movie Review dataset. Results of MLE, SeqGAN and LeakGAN are based on their published implementations. Since RankGAN didn't publish code, we cannot report the results of RankGAN on IMDB.}
\label{tab:avgBLEU-IMDB}
\end{table}

\begin{table*}[!t]\small\setlength{\tabcolsep}{3pt}
\centering
\begin{tabular}{c|c|c}
\toprule
Models&COCO&IMDB\\
\midrule
\tabincell{c p{0.09\textwidth}}{
MLE
}
&
\tabincell{p{0.4\textwidth}}{
(1) A girl sitting at a table in front of medical chair. \\
(2) The person looks at a bus stop while talking on a phone.
}
&
\tabincell{p{0.4\textwidth}}{
(1) If somebody that goes into a films and all the film cuts throughout the movie. \\
(2) Overall, it is what to expect to be she made the point where she came later.
}
\\
\midrule
\tabincell{c p{0.09\textwidth}}{
SeqGAN
}
&
\tabincell{p{0.4\textwidth}}{
(1) A man holding a tennis racket on a tennis court. \\
(2) A woman standing on a beach next to the ocean.
}
&
\tabincell{p{0.4\textwidth}}{
(1) The story is modeled after the old classic "B" science fiction movies we hate to love, but do. \\
(2) This does not star Kurt Russell, but rather allows him what amounts to an extended cameo.
}
\\
\midrule
\tabincell{c p{0.09\textwidth}}{
LeakGAN
}
&
\tabincell{p{0.4\textwidth}}{
(1) A bathroom with a toilet , window , and white sink. \\
(2) A man in a cowboy hat is milking a black cow.
}
&
\tabincell{p{0.4\textwidth}}{
(1) I was surprised to hear that he put up his own money to make this movie for the first time.  \\
(2) It was nice to see a sci-fi movie with a story in which you didn't know what was going to happen next.
}
\\
\midrule
\tabincell{c p{0.09\textwidth}}{
IRL\\(This work)
}
&
\tabincell{p{0.4\textwidth}}{
(1) A woman is standing underneath a kite on the sand.\\
(2) A dog owner walks on the beach holding surfboards.
}
&
\tabincell{p{0.4\textwidth}}{
(1) Need for Speed is a great movie with a very enjoyable storyline and a very talented cast.\\
(2) The effects are nothing spectacular, but are still above what you would expect, all things considered.
}
\\
\bottomrule
\end{tabular}
\caption{Case study. Generated texts from different models on COCO image caption and IMDB movie review datasets.}
\label{tab:generated_samples}
\end{table*}

\begin{table}[t]\setlength{\tabcolsep}{4pt}\small
\centering
\begin{tabular}{c|cccc||c} \toprule
Corpora & MLE & SeqGAN & LeakGAN & IRL & \tabincell{c}{Ground\\Truth} \\ \midrule
COCO & 0.205 & 0.450 & 0.543 & \textbf{0.550} & 0.725 \\
IMDB & 0.138 & 0.205 & 0.385 & \textbf{0.463} & 0.698 \\ \bottomrule
\end{tabular}
\caption{Results of Turing test. Samples of MLE, SeqGAN and LeakGAN are generated based on their published implementations. Since RankGAN didn't publish code, we cannot generate samples of RankGAN.}
\label{tab:Turing-test}
\end{table}

\subsection{IMDB Movie Reviews}
We use a large IMDB text corpus \cite{diao2014jointly} for training the generative models as long-length text generation. The dataset is a collection of 350K movie reviews. We select sentences with the length between 17 and 25, set word frequency at 180 as the threshold of frequently occurred words and remove sentences with low frequency words. Finally we randomly choose 80000 sentences for training and 3000 sentences for testing with the vocabulary size at 4979 and the average sentence length is 19.6.

IMDB is a more challenging corpus. Unlike sentences in COCO Image captions dataset, which mainly contains simple sentences, e.g., sentences only with the subject-predicate structure, IMDB movie reviews are comprised of various kinds of compound sentences. Besides, the sentence length of IMDB is much longer than that of COCO.

We also use the same metrics (BLEU$_\text{F}$, BLEU$_\text{B}$, BLEU$_\text{HA}$) to evaluate our method. 
The results in Table \ref{tab:avgBLEU-IMDB} show our method outperforms other models.


\subsection{Turing Test and Case Study}
The evaluation metrics mentioned above are still not sufficient for evaluating the quality of the sentences because they just focus on the local statistics, ignoring the long-term dependency characteristic of language. So we have to conduct a Turing Test based on scores by a group of people. Each sentence will get 1 point when it is viewed as a real one, otherwise 0 point. We perform the test on frameworks of MLE, SeqGAN, LeakGAN and our method on COCO Image captions dataset and IMDB movie review dataset.

Practically, we sample 20 sentences by each generator from different methods, and for each sentence, we ask 20 different people to score it. Finally, we compute the average score for each sentence, and then
calculate the average score for each method according to the sentences it generate.

Table \ref{tab:generated_samples} shows some generated samples of our and the baseline methods. These samples are what we have collected for people to score.

The results in Table \ref{tab:Turing-test} indicate that the generated sentences of our method have better quality than those generated by MLE, SeqGAN and LeakGAN, especially for long texts.

\section{Related Work}
Text generation is a crucial task in NLP which is widely used in a bunch of NLP applications. 
Text generation is more difficult than image generation since texts consist of sequential discrete decisions. Therefore, GAN fails to back propagate the gradients to update the generator. Recently, several methods have been proposed to alleviate this problem, such as Gumbel-softmax GAN \cite{kusner2016gans}, RankGAN \cite{lin2017adversarial}, TextGAN \cite{zhang2017adversarial}, LeakGAN \cite{guo2017long}, etc.

SeqGAN \cite{yu2017seqgan} addresses the differentiation problem by introducing RL methods, but still suffers from the problem of  reward sparsity. LeakGAN \cite{guo2017long} manages the reward sparsity problem via Hierarchical RL methods. \citet{toyama2018toward} designs several reward functions for partial sequence to solve the issue. However, the generated texts of these methods still lack diversity due to the mode collapse issue. In this paper, we employ IRL framework \cite{finn2016guided} for text generation. Benefiting from its inherent instant reward learning and entropy regularization, our method can generate more diverse texts.



\section{Conclusions \& Future Work}
In this paper, we propose a new method for text generation by using inverse reinforcement learning (IRL). This method alleviates the problems of reward sparsity and mode collapse in the adversarial generation models. In addition, we propose three new evaluation measures based on BLEU score to better evaluate the generated texts.

In the future, we would like to generalize the IRL framework to the other NLP tasks, such as machine translation, summarization, question answering, etc.

\section*{Acknowledgements}
We would like to thank the anonymous reviewers for their valuable comments. The research work is supported by the National Key Research and Development Program of China (No. 2017YFB1002104), Shanghai Municipal Science and Technology Commission (No. 17JC1404100 and 16JC1420401), and National Natural Science Foundation of China (No. 61672162).

\bibliographystyle{named}
\bibliography{ijcai18}

\end{document}